\documentclass[runningheads]{llncs}
\usepackage{graphicx}
\usepackage{tikz}
\usepackage{comment}
\usepackage{amsmath,amssymb} % define this before the line numbering.
\usepackage{color}
\usepackage{subfigure}
\usepackage[accsupp]{axessibility}  % Improves PDF readability for those with disabilities.

\begin{document}
% \linenumbers
\pagestyle{headings}
\mainmatter
\def\ECCVSubNumber{7}  % Insert your submission number here

\title{Perspective Reconstruction of Human Faces by Joint Mesh and Landmark Regression} % Replace with your title

%\begin{comment}
% \titlerunning{ECCV-22 submission ID \ECCVSubNumber} 
% \authorrunning{ECCV-22 submission ID \ECCVSubNumber} 
% \author{Anonymous ECCV submission}
% \institute{Paper ID \ECCVSubNumber}
%\end{comment}
%******************

% \begin{comment}
\titlerunning{Joint Mesh and Landmark Regression}
\author{Jia Guo\inst{1} \and
Jinke Yu\inst{1}  \and
Alexandros Lattas\inst{2} \and
Jiankang Deng\inst{1,2} 
}
\authorrunning{J. Guo et al.}
\institute{Insightface  \and
Imperial College London \\
\email{\{guojia, jackyu961127\}@gmail.com,\{a.lattas,j.deng16\}@imperial.ac.uk} 
}
% \end{comment}
%******************
\maketitle

\begin{abstract}
Even though 3D face reconstruction has achieved impressive progress, most orthogonal projection-based face reconstruction methods can not achieve accurate and consistent reconstruction results when the face is very close to the camera due to the distortion under the perspective projection. In this paper, we propose to simultaneously reconstruct 3D face mesh in the world space and predict 2D face landmarks on the image plane to address the problem of perspective 3D face reconstruction. Based on the predicted 3D vertices and 2D landmarks, the 6DoF (6 Degrees of Freedom) face pose can be easily estimated by the PnP solver to represent perspective projection. Our approach achieves 1st place on the leader-board of the ECCV 2022 WCPA challenge and our model is visually robust under different identities, expressions and poses. The training code and models are released to facilitate future research. \url{https://github.com/deepinsight/insightface/tree/master/reconstruction/jmlr}
\end{abstract}

\section{Introduction}

Monocular 3D face reconstruction has been widely applied in many fields such as VR/AR applications (e.g., movies, sports, games), video editing, and virtual avatars. Reconstructing human faces from monocular RGB data is a well-explored field and most of the approaches can be categorized into optimization-based methods \cite{Blanz1999,booth2018large,ganfit} or regression-based methods \cite{zhu20153ddfa,feng2018joint,deng2020retinaface,deng2019accurate}.

For the optimization-based methods, a prior of face shape and appearance \cite{Blanz1999,booth2018large} is used. The pioneer work 3D Morphable Model (3DMM) \cite{Blanz1999} represents the face shape and appearance in a PCA-based compact space and the fitting is then based on the principle of analysis-by-synthesis. In \cite{booth2018large}, an in-the-wild texture model is employed to greatly simplify the fitting procedure without optimization on the illumination parameters. To model textures in high fidelity, GANFit \cite{ganfit} harnesses Generative Adversarial Networks (GANs) to train a very powerful generator of facial texture in UV space and constrains the latent parameter by state-of-the-art face recognition model~\cite{deng2019arcface,arcface,deng2021variational,an2022killing,zhu2021webface260m,zhu2022webface260m}. However, the iterative optimization in these methods is not efficient for real-time inference.

For the regression-based methods, a series of methods are based on synthetic renderings of human faces \cite{genova2018unsupervised,wood20223d} or 3DMM fitted data \cite{zhu20153ddfa} to perform a supervised training of a regressor that predicts the latent representation of a prior face model (e.g., 3DMM~\cite{tran2016regressing}, GCN~\cite{zhou2019dense}, CNN~\cite{tran2018nonlinear}) or 3D vertices in different representation formats~\cite{jackson2017large,gueler2017densereg,feng2018joint,deng2020retinaface}. Genova et al.~\cite{genova2018unsupervised} propose a 3DMM parameter regression technique that is based on synthetic renderings and Tran et al.~\cite{tran2016regressing} directly regress 3DMM parameters using a CNN trained on fitted 3DMM data. Zhu et al.~\cite{zhu20153ddfa} propose 3D Dense Face Alignment (3DDFA) by taking advantage of Projected Normalized Coordinate Code (PNCC). In \cite{zhou2019dense}, joint shape and texture auto-encoder using direct mesh convolutions is proposed based on Graph Convolutional Network (GCN). 
In \cite{tran2018nonlinear}, CNN-based shape and texture decoders are trained on unwrapped UV space for non-linear 3D morphable face modelling. Jackson et al.~\cite{jackson2017large} propose a model-free approach that reconstructs a voxel-based representation of the human face. DenseReg~\cite{gueler2017densereg} regresses horizontal and vertical tessellation which is obtained by unwrapping the template shape and transferring it to the image domain.
PRN~\cite{feng2018joint} predicts a position map in the UV space of a template mesh. RetinaFace~\cite{deng2020retinaface} directly predicts projected vertices on the image plane within the face detector. Besides these supervised regression methods, there are also self-supervised regression approaches. Deng et al.~\cite{deng2019accurate} train a 3DMM parameter regressor based on photometric reconstruction loss with skin attention masks, a perception loss based on FaceNet~\cite{facenet}, and multi-image consistency losses. DECA~\cite{deca} robustly produces a UV displacement map from a low-dimensional latent representation.

Although the above studies have achieved good face reconstruction results, existing methods mainly use orthogonal projection to simplify the projection process of the face. When the face is close to the camera, the distortion caused by perspective projection can not be ignored \cite{kao2022single,zielonka2022towards}. Due to the use of orthogonal projection, existing methods can not well explain the facial distortion caused by perspective projection, leading to poor performance.

To this end, we propose a joint 3D mesh and 2D landmark regression method in this paper. Monocular 3D face reconstruction usually includes two tasks: 3D face geometric reconstruction and face pose estimation. However, we avoid explicit face pose estimation as in RetinaFace~\cite{deng2020retinaface} and DenseLandmark~\cite{wood20223d}. The insight behind this strategy is that the evaluation metric for face reconstruction is usually in the camera space and the regression error on 6DoF parameters will repeat to every vertex. Even though dense vertex or landmark regression seems redundant, the regression error of each point is individual. Most important, explicit pose parameter regression can result in drift-alignment problem \cite{chaudhuri2019joint}, which will make the 2D visualization of face reconstruction unsatisfying. 
By contrast, the 6 Degrees of Freedom (6DoF) face pose can be easily estimated by the PnP solver based on our predicted 3D vertices and 2D landmarks. 

To summarize, the main contributions of this work are:
\begin{itemize}
    \item We propose a Joint Mesh and Landmark Regression (JMLR) method to reconstruct 3D face shape under perspective projection, and the 6DoF face pose can be further estimated by PnP.
    \item The proposed JMLR achieves first place on the leader-board of the ECCV 2022 WCPA challenge.
    \item The visualization results show that the proposed JMLR is robust under different identities, exaggerated expressions and extreme poses.
\end{itemize}

\section{Our Method}

\subsection{3D Face Geometric Reconstruction}

\begin{figure}[t]
\centering
\includegraphics[width=1.0\linewidth]{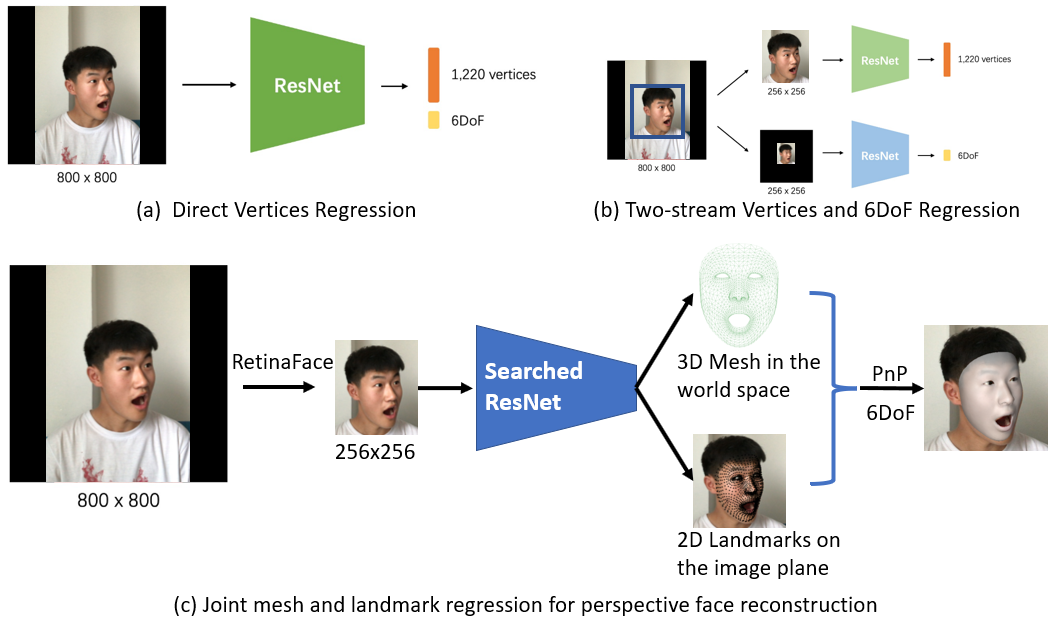}
\caption{Straightforward solutions for perspective face reconstruction. (a) direct 3D vertices regression, (b) two-stream 3D vertices regression and 6DoF prediction, and (c) our proposed Joint Mesh and Landmark Regression (JMLR).}
\label{fig:mesh}
\vspace{-4mm}
\end{figure}

In Fig.~\ref{fig:mesh}, we show three straightforward solutions for perspective face reconstruction. The simplest solution as shown in Fig.~\ref{fig:mesh} (a) is to directly regress the 3D vertices and 6DoF (i.e., Euler angles and translation vector) from the original image (i.e., $800\times800$), however, the performance of this method is very limited as mentioned by the challenge organizer. Another improved solution as illustrated in Fig.~\ref{fig:mesh} (b) is that the face shape should be predicted from the local facial region while the 6DoF can be obtained from the global image. Therefore, the local facial region is cropped and resized into $256\times256$, and then this face patch is fed into the ResNet \cite{he2016deep} to predict the 1,220 vertices. To predict 6DOF information, the region outside the face is blackened and the original $800\times800$ image is then resized into $256\times256$ as the input of another ResNet. 

In RetinaFace \cite{deng2020retinaface}, explicit pose estimation is avoided by direct mesh regression on the image plane as direct pose parameter regression can result in misalignment under challenging scenarios. However, RetinaFace only considered orthographic face reconstruction. In this paper, we slightly change the regression target of RetinaFace, but still employ the insight behind RetinaFace, that is, avoiding direct pose parameter regression. More specifically, we directly regress 3D facial mesh in the world space as well as projected 2D facial landmarks on the image plane as illustrated in Fig.~\ref{fig:mesh} (c).

For 3D facial mesh regression in the world space, we predict a fixed number of $N=1,220$ vertices ($ \mathbf{V} =[v_{x}^0, v_{y}^0, v_{z}^0; \cdots; v_{x}^{N-1}, v_{y}^{N-1}, v_{x}^{N-1}]$) on a pre-defined topological triangle context (i.e., 2,304 triangles). These corresponding vertices share the same semantic meaning across different faces. With the fixed triangle topology, every pixel on the face can be indexed by the barycentric coordinates and the triangle index, thus there exists a pixel-wise correspondence with the 3D face. In \cite{wood20223d}, 703 dense landmarks covering the entire head, including ears, eyes, and teeth, are proved to be effective and efficient to encode facial identity and subtle expressions for 3D face reconstruction. Therefore, 1K-level dense vertices/landmarks are a good balance between accuracy and efficiency for face reconstruction.

As each 3D face is represented by concatenating its $N$ vertex coordinates, we employ the following vertex loss to constrain the location of vertices:
\begin{eqnarray}\label{eq:verticsloss}
\mathcal{L}_{vert} {=} \frac{1}{N}\sum_{i=1}^{N} ||v_i(x,y,z)-v^{*}_i(x,y,z)||_{1},
\vspace{-4mm}
\end{eqnarray}
where $N=1,220$ is the number of vertices, $v$ is the prediction of our model and $v^{*}$ is the ground-truth.
By taking advantage of the 3D triangulation topology, we consider the edge length loss~\cite{deng2020retinaface}:
\begin{eqnarray}\label{eq:edgeloss}
\mathcal{L}_{edge} {=} \frac{1}{3M}\sum_{i=1}^{M} ||e_i-e^{*}_i||_{1},
\vspace{-4mm}
\end{eqnarray}
where $M=2,304$ is the number of triangles, $e$ is the edge length calculated from the prediction and $e^{*}$ is the edge length calculated from the ground truth. The edge graph is a fixed topology as shown in Fig.~\ref{fig:mesh} (c).

For projected 2D landmark regression, we also employ distance loss to constrain predicted landmarks close to the projected landmarks from ground-truth:
\begin{eqnarray}\label{eq:landloss}
\mathcal{L}_{land} {=} \frac{1}{N}\sum_{i=1}^{N} ||p_i(x,y)-p^{*}_i(x,y)||_{1},
\vspace{-4mm}
\end{eqnarray}
where $N=1,220$ is the number of vertices, $p$ is the prediction of our model and $p^{*}$ is the ground-truth generated by perspective projection.

By combining the vertex loss and the edge loss for 3D mesh regression and the landmark loss for 2D projected landmark regression, we define the following prospective face reconstruction loss:
\begin{eqnarray}\label{eq:totalloss}
\mathcal{L} {=}  \mathcal{L}_{vert} + \lambda_0 \mathcal{L}_{edge} + \lambda_1 \mathcal{L}_{land},
\vspace{-4mm}
\end{eqnarray}
where $\lambda_0$ is set to $0.25$ and $\lambda_1$ is set to $2$ according to our experimental experience.

\subsection{6DoF Estimation}

Based on the predicted 3D vertices in the world space, projected 2D landmarks on the image plane, and the camera intrinsic parameters, we can easily employ the Perspective-n-Point (PnP) algorithm \cite{marchand2015pose} to compute the 6D facial pose parameters (i.e., the rotation of roll, pitch, and yaw as well as the 3D translation of the camera with respect to the world).
Even though directly regressing 6DoF pose parameters from a single image by CNN (Fig.~\ref{fig:mesh} (a)) is also feasible, it achieves much worse performance than our
method due to the non-linearity of the rotation space. 

In perspective projection, 3D face shape $V_{world}$ is transformed from the world coordinate system to the camera coordinate system by using 6DoF face pose (i.e., the rotation matrix $R\in \mathbb{R}^{3\times3}$ and the translation vector $T\in \mathbb{R}^{1\times3}$) with known intrinsic camera parameters $K$,
\begin{eqnarray}\label{eq:pnp}
V_{camera} = K (V_{world}R+T),
\vspace{-4mm}
\end{eqnarray}
where $K$ is related to (1) the coordinates of the principal point (the intersection of the optical axes with the image plane) and (2) the ratio between the focal length and the size of the pixel. The intrinsic parameters $K$ can be easily obtained through an
off-line calibration step. Knowing 2D-3D point correspondences as well as the intrinsic parameters, pose estimation is straightforward by calling \emph{cv.solvePnP().}

\section{Experimental Results}

\subsection{Dataset}
In the perspective face reconstruction challenge, 
250 volunteers were invited to record the training and test dataset. These volunteers sit in a random environment, and the 3D acquisition equipment (i.e., iPhone 11) is fixed in front of them, with a distance ranging from about 0.3 to 0.9 meters. Each subject is asked to perform 33 specific expressions with two head movements (from looking left to looking right / from looking up to looking down). The triangle mesh and head pose information of the RGB image is obtained by the built-in ARKit toolbox. Then, the original data are pre-processed to unify the image size and camera intrinsic, as well as eliminate the shifting of the principal point of the camera.

In this challenge, 200 subjects with 356,640 instances are used as the training set, and 50 subjects with 90,545 instances are used as the test set. Note that there is only one face in each image and the location of each face is provided by the face-alignment toolkit \cite{bulat2017far}.
The ground truth of 3D vertices and pose transform matrix of the training set is provided. The unit of 3D ground-truth is in meters. The facial triangle mesh is made up of 1,220 vertices and 2,304 triangles. The indices of 68 landmarks \cite{deng2019menpo} from 1,220 vertices are also provided.

\subsection{Evaluation Metrics}

For each test image, the challenger should predict the 1,220 3D vertices in world space (i.e., $V_{world}\in \mathbb{R}^{1220\times3}$) and the pose transform matrix (i.e., the rotation matrix $R\in \mathbb{R}^{3\times3}$ and the translation vector $T\in \mathbb{R}^{1\times3}$) from world space to camera space.

$V_{world}=
\begin{bmatrix}
v_x^0 & v_y^0 & v_z^0  \\ 
v_x^1 & v_y^1 & v_z^1  \\ 
 \vdots & \vdots  & \vdots  \\ 
v_x^{1219} & v_y^{1219} & v_z^{1219}  
\end{bmatrix}$, 
$R=\begin{bmatrix}
r_{00} & r_{01}  & r_{02}   \\ 
r_{10}  & r_{11}  & r_{12}   \\  
r_{20}  & r_{21}  & r_{22}   
\end{bmatrix}$,
$T = \begin{bmatrix}
t_x & t_y & t_z
\end{bmatrix}$.

Then, we can compute the transformed vertices in camera space by $V_{camera} = V_{world}R+T$. 
In this challenge, the error of 3D vertices and face pose is measured in camera space. Four transformed sets of vertices are computed as follows:
\begin{flalign}
V_1 & = V^{gt}R^{gt}+T^{gt},  
V_2  = V^{pred}R^{pred}+T^{pred},\\ \nonumber
V_3 & = V^{gt}R^{pred}+T^{pred}, 
V_4  = V^{pred}R^{gt}+T^{gt}.
\end{flalign}

Finally, $L_2$ distance between pair $(V_1, V_2), (V_1, V_3), (V_1, V_4)$ are calculated and combined into the final distance error:
\begin{equation}
L_{error}=\left \|V_1-V_2  \right \|_2 +\left \|V_1-V_3  \right \|_2 + 10 \left \|V_1-V_4  \right \|_2,
\end{equation}
where geometry accuracy across different identities and expressions is emphasized by $\times 10$.
On the challenge leader-board, the distance error is multiplied by 1,000, thus the distance error is in millimeters instead of meters.

\subsection{Implementation Details}
\noindent{\bf Input.} We employ five facial landmarks predicted by RetinaFace \cite{deng2020retinaface} for normalized face cropping at the resolution of $256 \times 256$. Colour jitter and flip data augmentation are also used during our training. 

\begin{figure}
\centering
\subfigure[ResNet 34]{
\label{fig:backbonebasline}
\includegraphics[height=0.3\textwidth]{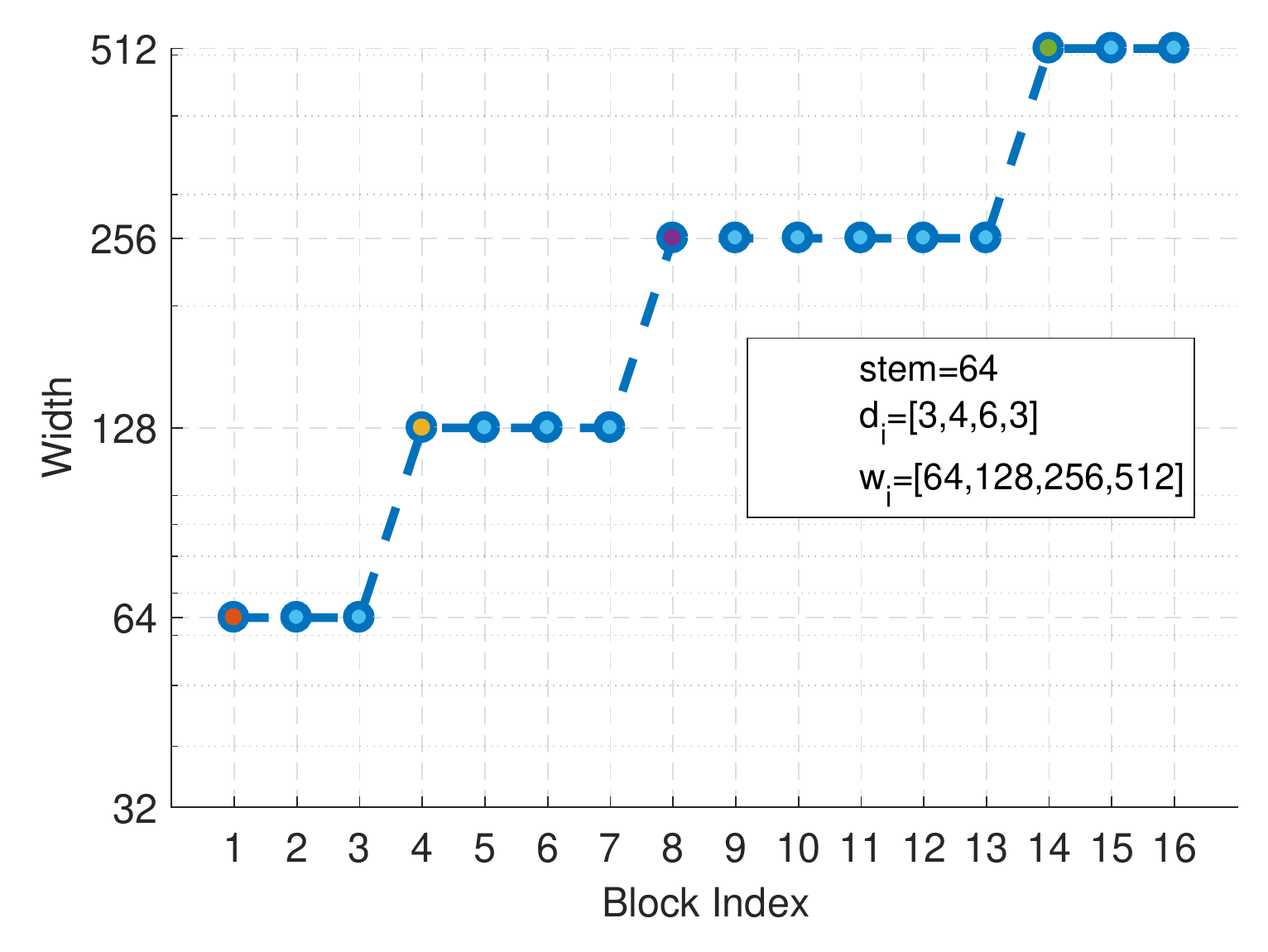}}
\subfigure[Searched ResNet]{
\label{fig:searchedresnet}
\includegraphics[height=0.3\textwidth]{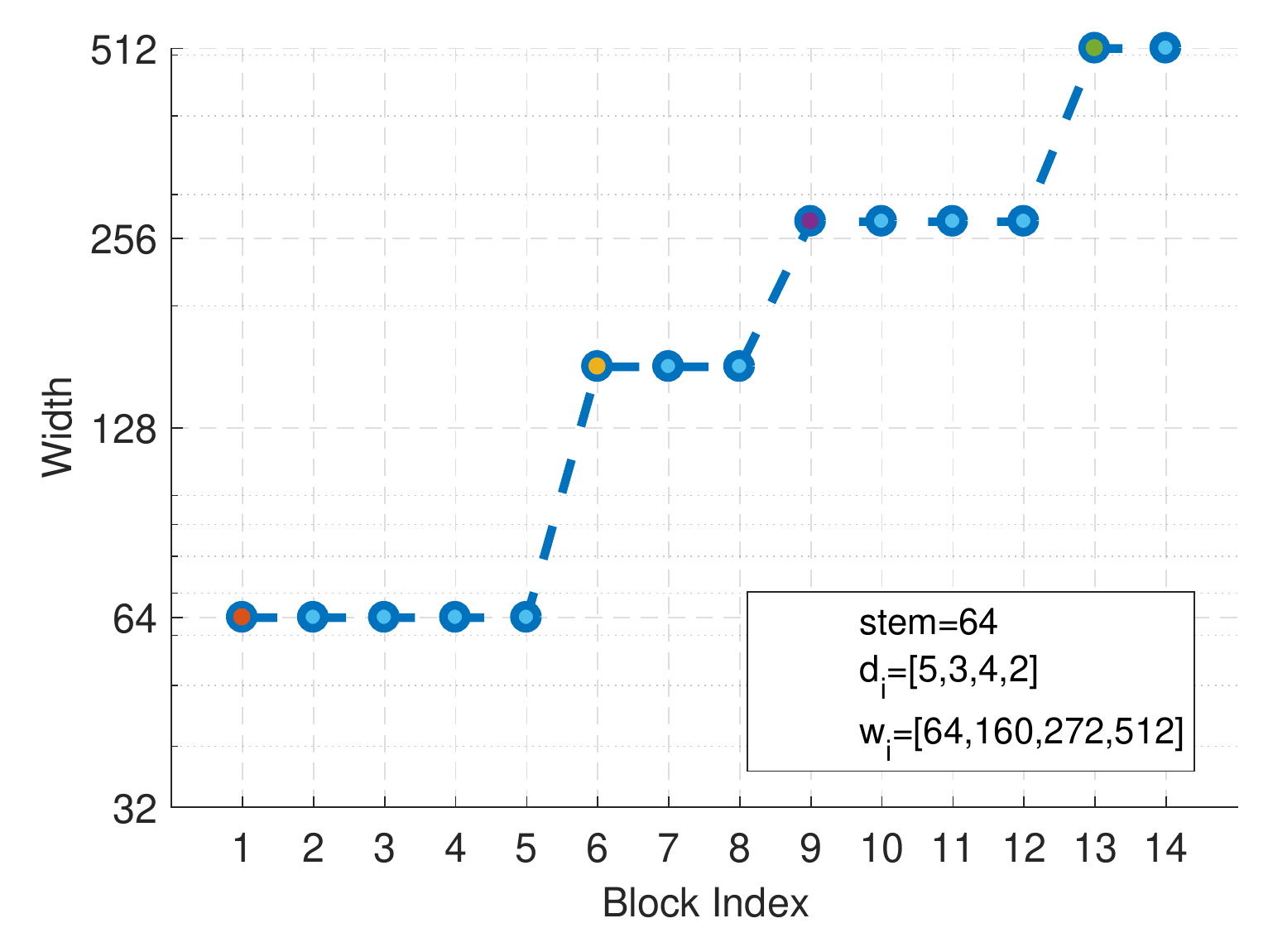}}
\vspace{-2mm}
\caption{Computation redistribution for the backbone. (a) the default ResNet-34 design. (b) our searched ResNet structure for perspective face reconstruction.}
\vspace{-4mm}
\label{fig:backbone}
\end{figure}

\noindent{\bf Backbone.} We employ the ResNet \cite{he2016deep} as our backbone. In addition, we refer to SCRFD \cite{guo2021sample} to optimize the computation distribution over the backbone. For the ResNet design, there are four stages (i.e., C2, C3, C4 and C5) operating at progressively reduced resolution, with each stage consisting of a sequence of identical blocks. For each stage $i$, the degrees of freedom include the number of blocks $d_i$ (i.e., network depth) and the block width $w_i$ (i.e, number of channels). Therefore, the backbone search space has 8 degrees of freedom as there are 4 stages and each stage $i$ has 2 parameters: the number of blocks $d_i$ and block width $w_i$. Following RegNet \cite{radosavovic2020designing}, we perform uniform sampling of $d_i\leq 24$ and $w_i\leq 512$ ($w_i$ is divisible by 8). As state-of-the-art backbones have increasing widths, we also constrain the search space, according to the principle of $w_{i+1}\ge w_i$. The searched ResNet structure for perspective face reconstruction is illustrated in Fig~\ref{fig:searchedresnet}. 

\noindent{\bf Loss optimization.} We train the proposed joint mesh and landmark regression method for 40 epochs with the SGD optimizer. We set the learning rate as $0.1$ and the batch size is set to $64\times 8$. The training is warmed up by 1,000 steps and then the Poly learning scheduler is used.

\subsection{Ablation Study}
In Table~\ref{table_ablationstudy}, we run several experiments to validate the effectiveness of the proposed augmentation, network structure, and loss. We split the training data into 80\% subjects for training and 20\% subjects for validation. From Table~\ref{table_ablationstudy}, we have the following observations: (1) flipping augmentation can decrease the reconstruction error by 0.45 as we check experiments 1 and 2, (2) the capacity of ResNet-34 is well matched to the dataset as we compare the performance between different network structures (e.g., ResNet-18, ResNet-34, and ResNet-50), (3) searched ResNet with similar computation cost as ResNet-34 can slightly decrease the reconstruction error by 0.18, (4) without the topology regularization by the proposed edge loss, the reconstruction error increase by 0.29, confirming the effectiveness of mesh regression proposed in RetinaFace \cite{deng2020retinaface}, (5) by the model ensemble, the reconstruction error significantly reduces by 0.92. 

\begin{table}[t]
\centering
\caption{Ablation study regarding augmentation, network structure, and loss. The results are reported on 40 subjects from the training set and the rest 160 subjects are used for training.} \label{table_ablationstudy}
	
\begin{tabular}{|c|c|c|c|c|}
\hline
  Aug& Network & Loss & Flops & Score \\
\hline
 Flip    & ResNet34 & $\mathcal{L}$ & 5.12G &  34.31 \\
 No Flip & ResNet34 & $\mathcal{L}$ & 5.12G &  34.76 \\
 Flip & ResNet18 & $\mathcal{L}$ & 3.42G &  34.95 \\
 Flip & ResNet50 & $\mathcal{L}$ & 5.71G & 34.70 \\
 Flip & Searched ResNet & $\mathcal{L}$ & 5.13G &  34.13 \\ 
 Flip & ResNet34 & $\mathcal{L}_{vert}$+$\mathcal{L}_{land}$  & 5.12G &  34.60 \\
 Flip & ResNet34 \& Searched ResNet & $\mathcal{L}$ & 10.25 G & 33.39 \\
\hline
\end{tabular}
\vspace{-2mm}
\end{table}

\subsection{Benchmark Results}
For the submission of the challenge, we employ all of the 200 training subjects and combine all the above-mentioned training tricks (e.g. flip augmentation, searched ResNet, joint mesh and landmark regression, and model ensemble). As shown in Table~\ref{table_benchmark}, we rank first on the challenge leader-board and our method outperforms the runner-up by 0.07. 

\begin{table}[t]
\centering
\caption{Leaderboard results. The proposed JMLR achieves best overall score compared to other methods.} \label{table_benchmark}
\begin{tabular}{|l|c|c|c|c|c|}
\hline
Rank & Team & Score &  $ \left \|V_1-V_2  \right \|_2$ & $\left \|V_1-V_3  \right \|_2$ & $10 \left \|V_1-V_4  \right \|_2$\\
\hline
1st (JMLR)& EldenRing  &\textbf{32.811318} & \textbf{8.596190} & \textbf{8.662822} & 15.552306\\
2nd & faceavata        & 32.884849         & 8.833569 & 8.876343 & \textbf{15.174936}\\
3rd & raccoon\&bird    & 33.841992         & 8.880053 & 8.937624 & 16.024315\\
\hline
\end{tabular}
\vspace{-4mm}
\end{table}

\subsection{Result Visualization}
\begin{figure}
\centering
\includegraphics[width=1.0\textwidth]{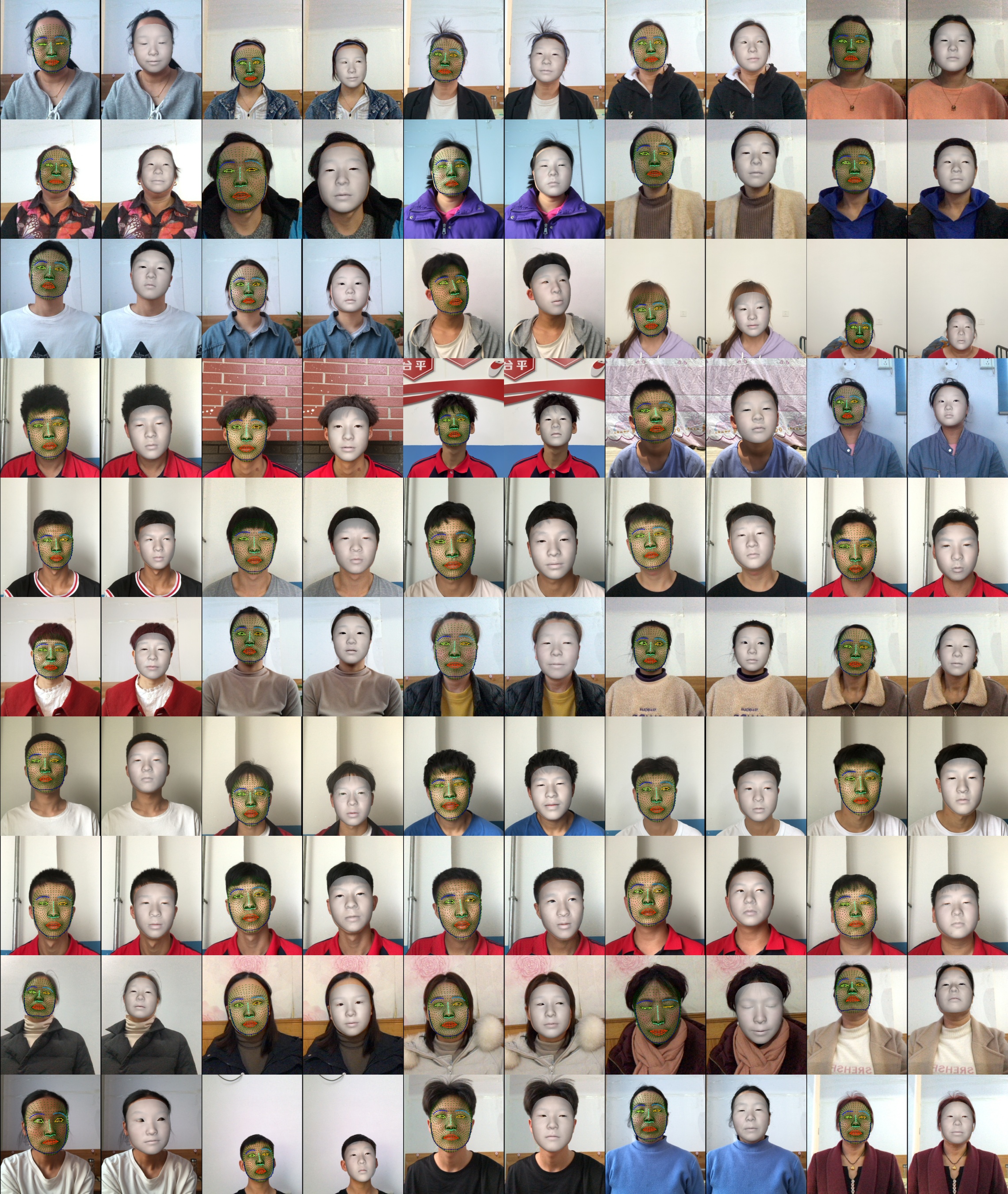} 
\caption{Predicted meshes on 50 test subjects. 68 landmarks are also indexed for better visualization. The proposed method shows stable and accurate mesh regression results across different subjects (i.e., genders and ages).}
\label{img_id}
\end{figure}

\begin{figure}
\centering
\includegraphics[width=1.0\textwidth]{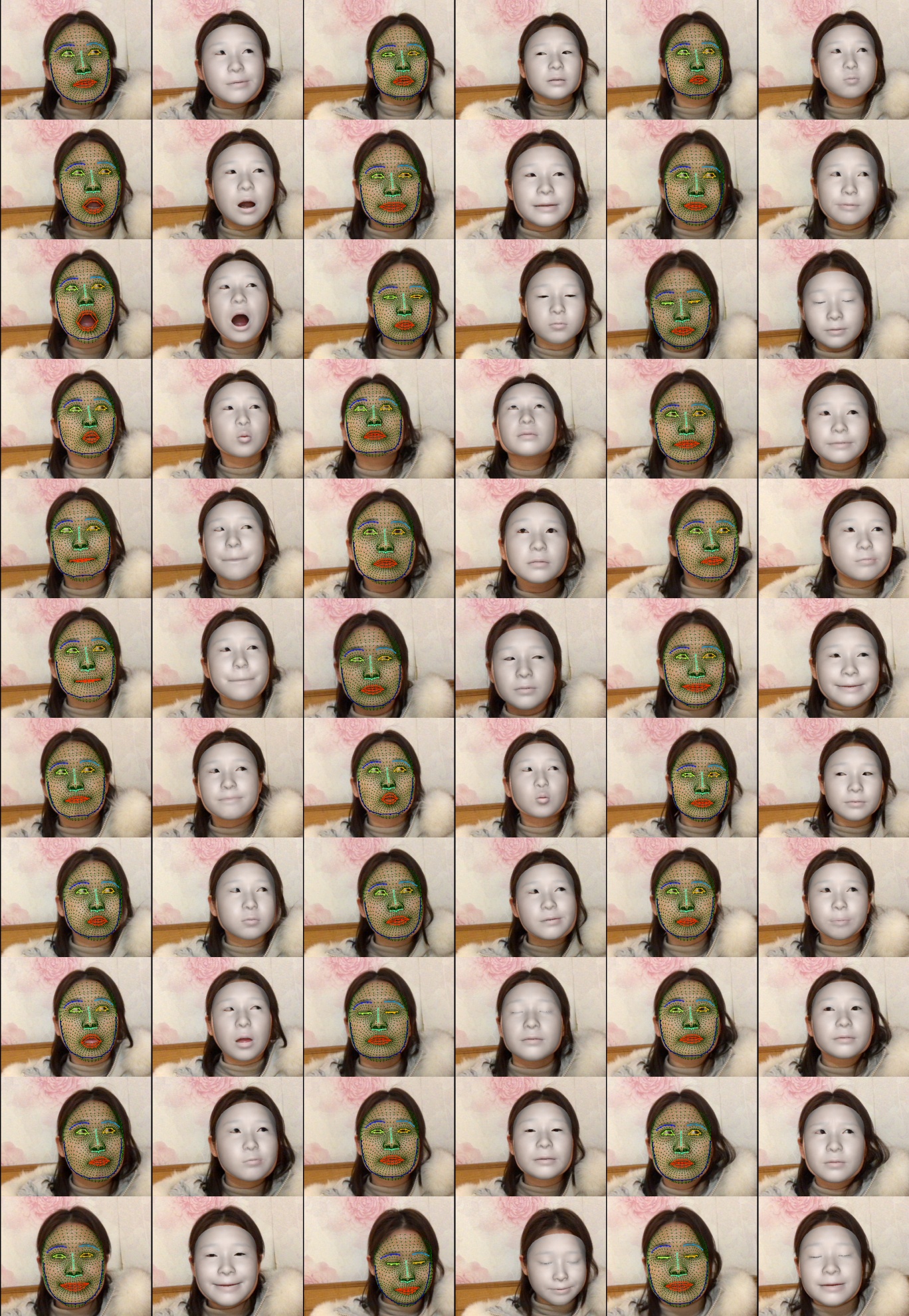} 
\caption{Predicted meshes of ID-6159 under 33 different expressions. 68 landmarks are also indexed for better visualization. The proposed method shows stable and accurate mesh regression results across different expressions (i.e., blinking and mouth opening).}
\label{img_exp6159}
\end{figure}

\begin{figure}
\centering
\includegraphics[width=1.0\textwidth]{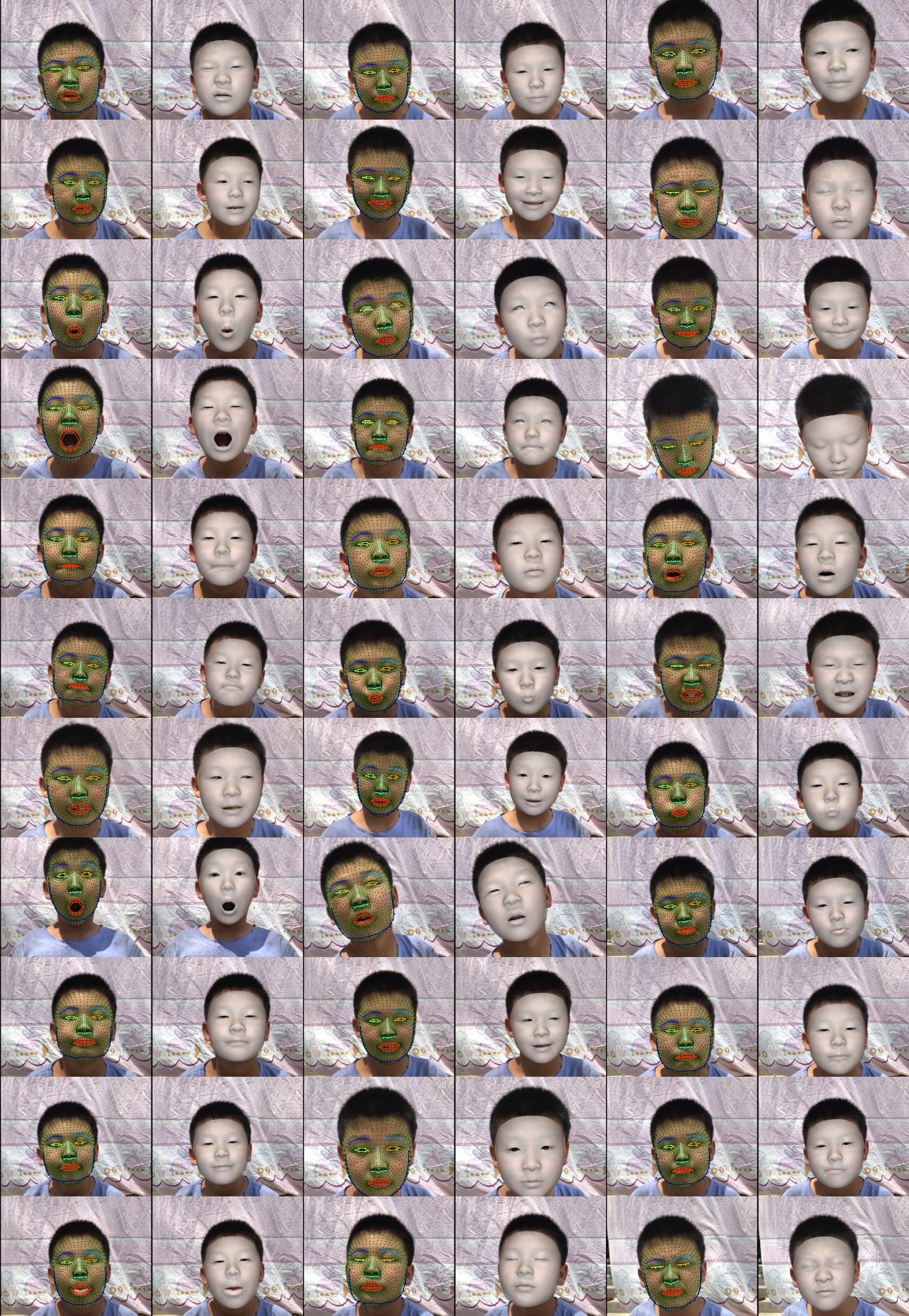} 
\caption{Predicted meshes of ID-239749 under 33 different expressions. 68 landmarks are also indexed for better visualization. The proposed method shows stable and accurate mesh regression results across different expressions (i.e., blinking and mouth opening).}
\label{img_exp239749}
\end{figure}

\begin{figure}
\centering
\includegraphics[width=1.0\textwidth]{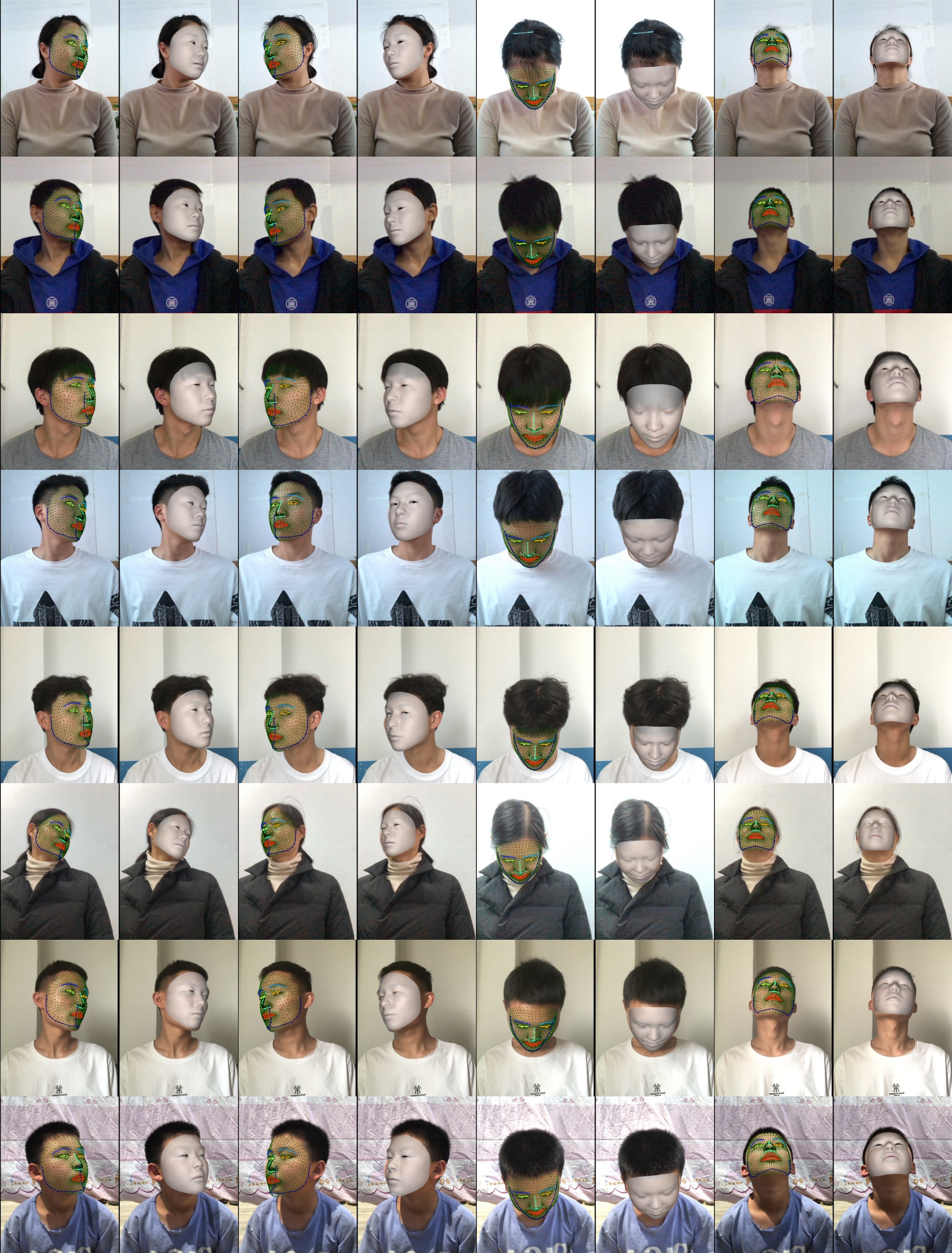} 
\caption{Predicted meshes of different identities under different poses. 68 landmarks are also indexed for better visualization. The proposed method shows stable and accurate mesh regression results across different facial poses (i.e., yaw and pitch).}
\label{img_pose1}
\end{figure}

\begin{figure}
\centering
\includegraphics[width=1.0\textwidth]{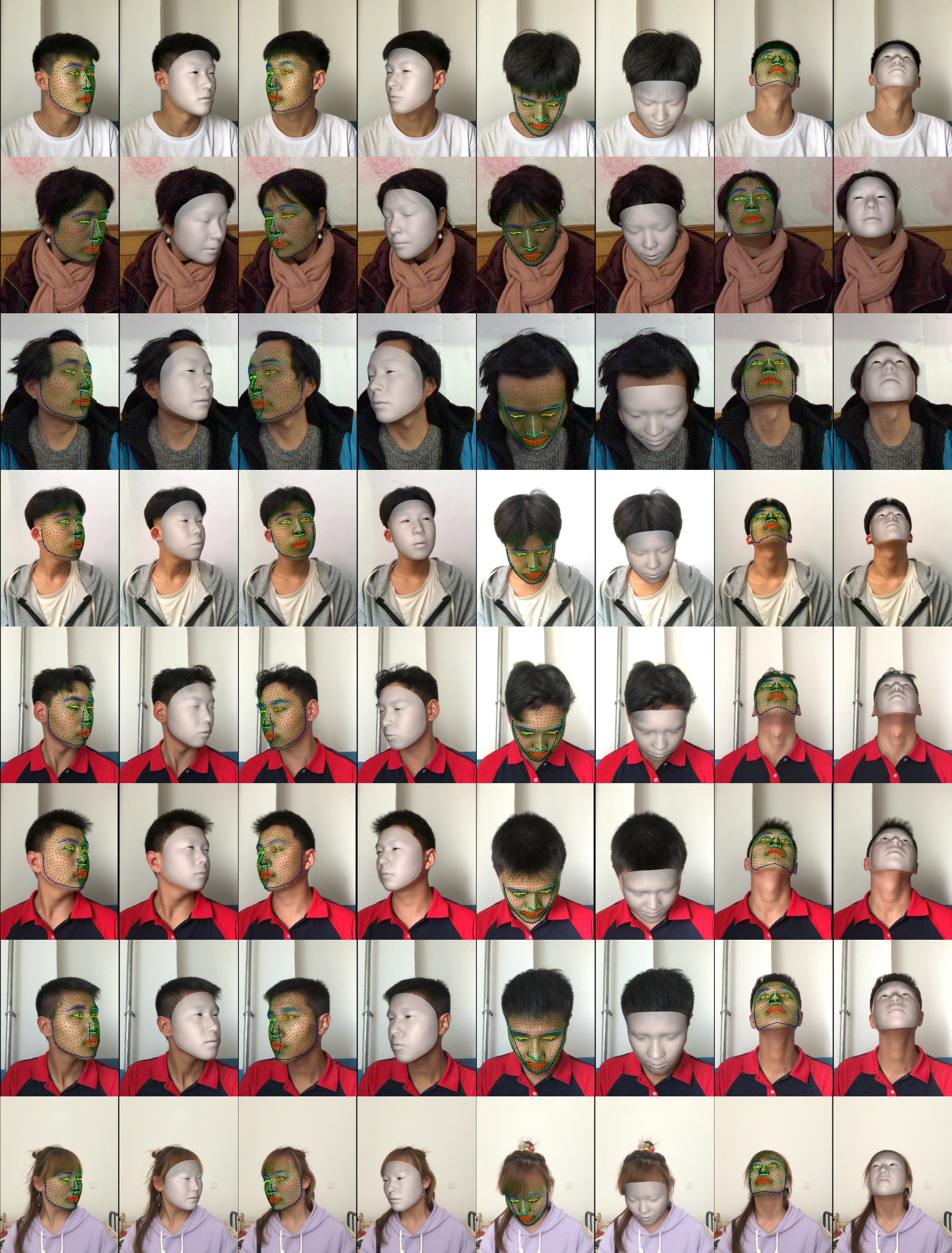} 
\caption{Predicted meshes of different identities under different poses. 68 landmarks are also indexed for better visualization. The proposed method shows stable and accurate mesh regression results across different facial poses (i.e., yaw and pitch).}
\label{img_pose3}
\end{figure}

\begin{figure}
\centering
\includegraphics[width=1.0\textwidth]{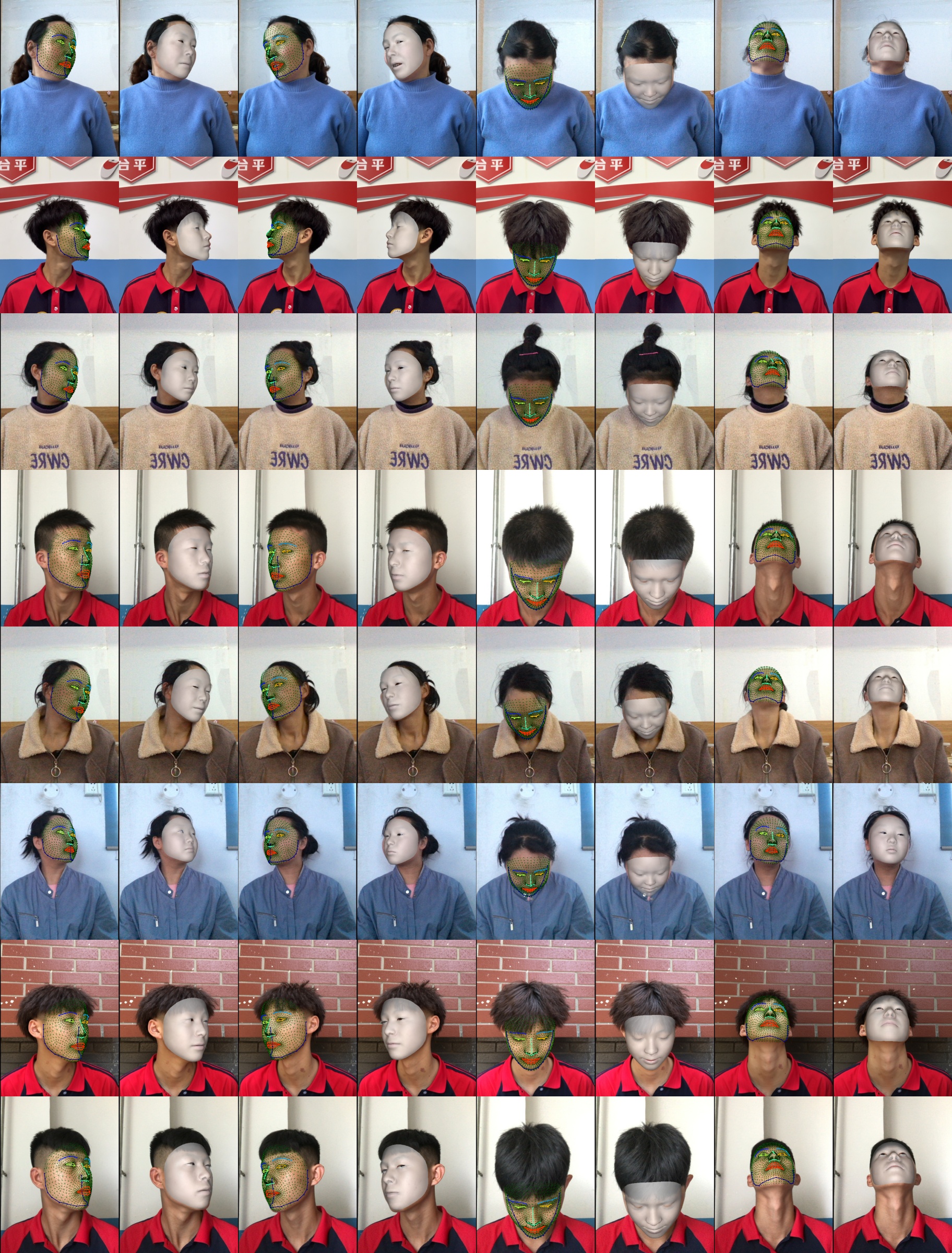} 
\caption{Predicted meshes of different identities under different poses. 68 landmarks are also indexed for better visualization. The proposed method shows stable and accurate mesh regression results across different facial poses (i.e., yaw and pitch).}
\label{img_pose4}
\end{figure}

In this section, we display qualitative results for perspective face reconstruction on the test dataset. As we care about the accuracy of geometric face reconstruction and 6DoF estimation, we visualize the projected 3D vertices on the image space both in the formats of vertex mesh and rendered mesh. As shown in Fig.~\ref{img_id}, the proposed method shows accurate mesh regression results across 50 different test subjects (i.e., genders and ages). In Fig.~\ref{img_exp6159} and Fig.~\ref{img_exp239749}, we select two subjects under 33 different expressions. The proposed method demonstrates precise mesh regression results across different expressions (i.e., blinking and mouth opening). In Fig.~\ref{img_pose1}, Fig.~\ref{img_pose3}, and Fig.~\ref{img_pose4}, we show the mesh regression results under extreme poses (e.g., large yaw and pitch variations). The proposed method can easily handle profile cases as well as large pitch angles. From these visualization results, we can see that our method is effective under different identities, expressions and large poses. 

% \begin{figure}
% \centering
% \includegraphics[width=1.0\textwidth]{figure/6068exp.jpg} 
% \caption{Predicted meshes of ID-6068 under 33 different expressions. 68 landmarks are also indexed out for better visualization. The proposed method shows stable and accurate mesh regression results across different expressions (i.e., blinking and mouth opening).}
% \label{img_exp6068}
% \end{figure}

% \begin{figure}
% \centering
% \includegraphics[width=1.0\textwidth]{figure/236848exp.jpg} 
% \caption{Predicted meshes of ID-236848 under 33 different expressions. 68 landmarks are also indexed out for better visualization. The proposed method shows stable and accurate mesh regression results across different expressions (i.e., blinking and mouth opening).}
% \label{img_exp236848}
% \end{figure}

% \begin{figure}
% \centering
% \includegraphics[width=1.0\textwidth]{figure/pose2.jpg} 
% \caption{Predicted meshes of different identities under different poses. 68 landmarks are also indexed out for better visualization. The proposed method shows stable and accurate mesh regression results across different facial poses (i.e., yaw and pitch).}
% \label{img_pose2}
% \end{figure}

\section{Conclusion}
In this paper, we explore 3D face reconstruction under perspective projection from a single RGB image. We implement a straightforward algorithm, in which joint face 3D mesh and 2D landmark regression are proposed for perspective 3D face reconstruction. We avoid explicit 6DoF prediction but employ a PnP solver for 6DoF face pose estimation given the predicted 3D vertices in the world space and predicted 2D landmarks on the image space. Both quantitative and qualitative experimental results demonstrate the effectiveness of our approach for perspective 3D face reconstruction and 6DoF pose estimation. Our submission to the ECCV 2022 WCPA challenge ranks first on the leader-board and the training code and pre-trained models are released to facilitate future research in this direction.

\clearpage
\bibliographystyle{splncs04}
\bibliography{egbib}
\end{document}